\documentclass[preprint,12pt]{elsarticle}

\usepackage[english]{babel}
\usepackage{natbib}

\usepackage{enumerate}
\usepackage{subcaption}
\usepackage{dirtytalk}
\usepackage{array}
\usepackage{xspace}
\usepackage{tipa}
\usepackage{mathtools}
\usepackage{graphicx}
\usepackage{graphbox} 
\usepackage{amssymb}
\usepackage{amsmath}
\usepackage{algorithm}
\usepackage[noend]{algpseudocode}
\usepackage[acronym]{glossaries}
\usepackage[utf8]{inputenc}
\usepackage[T1]{fontenc}
\usepackage{nomencl}
\usepackage[cal=dutchcal]{mathalfa}
\usepackage{pifont}
\usepackage{parskip}
\usepackage{caption,subcaption}
\usepackage{amsthm}
\theoremstyle{definition}
\newtheorem{definition}{Definition}[section]
\newtheorem{corollary}{Corollary}[section]
\newtheorem{theorem}{Theorem}[section]

\DeclareMathAlphabet{\mathpzc}{T1}{pzc}{m}{it}
\graphicspath{ {./images/} }

\journal{Journal of Manufacturing Systems}

\begin{document}

\begin{frontmatter}



\title{Graph neural networks-based Scheduler for Production planning problems using Reinforcement Learning}

\author[inst1]{Mohammed Sharafath Abdul Hameed}
\author[inst1]{Andreas Schwung}

\affiliation[inst1]{organization={Department of Automation Technology and Learning Systems}, addressline={South Westphalia University of Applied Sciences}, 
          city={Soest},
          postcode={59494}, 
          country={Germany}}

\begin{abstract}
Reinforcement learning (RL) is increasingly adopted in job shop scheduling problems (JSSP). But RL for JSSP is usually done using a vectorized representation of machine features as the state space. It has three major problems: (1) the relationship between the machine units and the job sequence is not fully captured, (2) exponential increase in the size of the state space with increasing machines/jobs, and (3) the generalization of the agent to unseen scenarios. This paper presents a novel framework named GraSP-RL, GRAph neural network-based Scheduler for Production planning problems using Reinforcement Learning. It represents JSSP as a graph and trains the RL agent using features extracted using a graph neural network (GNN). While the graph is itself in the non-euclidean space, the features extracted using the GNNs provide a rich encoding of the current production state in the euclidean space. At its core is a custom message-passing algorithm applied to the GNN. The node features encoded by the GNN are then used by the RL agent to select the next job. Further, we cast the scheduling problem as a decentralized optimization problem in which the learning agent is assigned to all the production units individually and the agent learns asynchronously from the experience collected on all the other production units. The GraSP-RL is then applied to a complex injection molding production environment with 30 jobs and 4 machines. The task is to minimize the makespan of the production plan. The schedule planned by GraSP-RL is then compared and analyzed with a priority dispatch rule algorithm like first-in-first-out (FIFO) and metaheuristics like tabu search (TS) and genetic algorithm (GA). The proposed GraSP-RL outperforms the FIFO, TS, and GA for the trained task of planning 30 jobs in JSSP. We further test the generalization capability of the trained agent on two different problem classes: Open shop system (OSS) and Reactive JSSP (RJSSP). In these modified problem classes our method produces results better than FIFO and comparable results to TS and GA, without any further training while also providing schedules instantly.
\end{abstract}

\begin{highlights}
\item Job Shop Scheduling
\item Reinforcement Learning
\item Graph Neural Network
\item Production Planning
\end{highlights}


\end{frontmatter}


\section{Introduction}
\label{sec:intro}

The task of manufacturing or production scheduling receives increasing attention from manufacturing enterprises to increase profitability and productivity on the shop floor, especially in a globally competitive market. The Job Shop Scheduling Problem (JSSP) is concerned with the allocation of tasks within a stipulated time to a limited number of resources in such a way that one or more objectives are optimized, for e.g.: lower production costs, lower setup time, shorter planned time, etc. The objectives are met by deriving an appropriate time to execute each operation by taking into account temporal relations between production processes and constraints of the shared manufacturing resources. JSSP is considered an NP-hard problem. 

The majority of approaches for JSSP are implemented as centralized algorithms which have full knowledge of the production process, like heuristics~\cite{LUDOF.GELDERS.1978}, constraint propagation and satisfaction~\cite{Baptiste.1995}, petri-nets~\cite{Lee.1994}, neural networks~\cite{Jain.1998}, expert systems~\cite{Kanet.1987} and metaheuristics~\cite{Mhasawade2017}. However, such centralized approaches have a few inherent drawbacks. They do not explicitly model the relationship between the machines and they also rely on a single computing unit which makes large-scale environments difficult as the problem size increases exponentially.

Furthermore, the optimization has to be repeated when the information in the production environment changes, i.e the above architectures also do not capture transferable knowledge. The knowledge capture problem can be overcome by using a reinforcement learning approach. But the solutions using reinforcement learning usually depend on a vectorized representation of the production state~\cite{Zhang.1995}~\cite{Gabel.2008}, which is also not scalable linearly.

We solve the above drawbacks in GraSP-RL by 1) modeling the production environment as a graph to explicitly capture the internal relationship between the machines and the buffers, 2) using a custom graph neural network (GNN) to extract rich feature encoding of a machine node, 3) using distributed learning on each machine node rather than the entire production unit, 4) using only the local features of the machine for the distributed agent. The production plant is modeled as a graph, where we develop a bipartite graph structure between the machine and the machine buffers. This provides a way to handle the complex interaction between the jobs in various machines and the buffers of others machines while abstracting away low-level relations. This is captured through an algorithm called message passing in the GNNs. Next, a reinforcement learning (RL) agent operates on the encoded, message-passed node features of the current machine to take action. Thus, the global optimization problem in a production unit is decomposed into a considerably smaller local optimization problem at the machine level for the RL agent. The optimization problem is thus made linearly scalable as RL agents can be added or removed as required into the graph-structured JSSP even after training.

The result is an RL agent that has exposure to multiple scenarios arising from just the locally seen information. An agent that is trained like so, generalizes well to unseen scenarios as GNNs are universally invariant to the order of the nodes~\cite{keriven2019universal} and their ability to be powerful injective mapping functions for non-isomorphic graphs~\cite{xu2018powerful}. This allows for a scalable architecture and high adaptability in large-scale dynamic production environments. 

The contributions of the work can be summarized as follows:
\begin{itemize}
	\item We develop a novel framework, GraSP-RL, to model the production environment as a bipartite graph that enables information transfer between the machines and machine buffers. The production graph is then processed by a GNN to extract features that are further used by a distributed RL agent to take action. The integration of GNN with the RL allows for an end-to-end trainable framework for the production scheduling problem.
	\item The distributed nature of the learning reduces the computation complexity of the neural network in the agent to be independent of the number of machines in the production environment enabling GraSP-RL to be scalable after training without additional computation costs.
	\item We apply GraSP-RL to a complex injection molding production planning benchmark problem with 30 jobs and 4 machines and compare the results with dispatch rule algorithms like first-in-first-out (FIFO) and metaheuristic algorithms like tabu search (TS) and genetic algorithm (GA). The results show that the RL agent trained with GraSP-RL performs better than FIFO, TS, and GA.
	\item We test the generalization capability of the RL agent by applying two perturbations to the JSSP environment, an open shop system (OSS) and reactive JSSP (RJSSP). GraSP-RL produces better results than FIFO and comparable results to TS and GA, without any training further reducing the computation costs.
\end{itemize}

The paper is organized as follows. Related work is presented in Section~\ref{sec:relwork}. In Section~\ref{sec:Prelims}, all the preliminary concepts are introduced that will be later used in GraSP-RL. Section~\ref{sec:jsspgraph}, introduces GraSP-RL theoretically and explains its workings. Further, Section~\ref{sec:expres}, presents the environment in which GraSP-RL is applied while providing empirical results, including ablation, comparative, and generalization studies. Finally, Section~\ref{sec:conclusion}, concludes the paper and provides the future outlook.

\section{Related Work}\label{sec:relwork}

Scheduling problems are known to be NP-hard~\cite{J.K.Lenstra.1979}, meaning it is considered that they cannot be solved in polynomial time. While early on the focus was on providing an exact solution, the realization that the scheduling problems are NP-hard redirected the research focus to approximate solutions. And as such the methods to solve scheduling problems are divided into two: exact and approximate methods~\cite{otala2021}. While exact methods include integer programming~\cite{Ku.2016}~\cite{Ozguven.2012} and branch and bound~\cite{hariri1991}, approximate methods include genetic algorithm~\cite{Fang.1993}, and tabu search~\cite{GLOVER1986533}~\cite{ponnambalam2000}~\cite{Baykasoglu.2004}. Although approximate methods provide a good solution in a reasonable time, they suffer from the ‘curse of dimensionality’, whereby the problem size increases exponentially through the addition of machines or jobs to the problem. And also there is no knowledge captured in these algorithms that can be reused at another time, i.e., these algorithms make a plan based on the initial condition and if those conditions change then the algorithm has to run again to find the near-optimal plan. In this work, we compare GraSP-RL, to two prominent metaheuristics methods: TS and GA. 

Recent improvements in RL have opened up a very interesting avenue to not only create a near-optimal plan but could also enable the capture of the planning knowledge such that it can be reused. RL has achieved super-human performance in playing board games such as Go, Chess~\cite{Silver.2018}~\cite{Schulman.2017}~\cite{Silver.2017}~\cite{Schrittwieser.2020} and robotics~\cite{Gu.2017}. Not only is RL capable of learning and performing a specific task well, but it can also generalize well~\cite{Mnih.2015} and has proven to be capable of multi-task learning~\cite{hessel2019}. The advancements in the capabilities of deep neural networks have also helped push the boundaries in the application of RL~\cite{Liu.2017}. \cite{Zhang.1995} was one of the first works to use RL to solve a static scheduling problem. Further,~\cite{Gabel.2008} used a model-free multi-agent approach to develop an adaptive reactive scheduling agent. Although these works still suffer from the curse of dimensionality it can be mitigated to some extent by using a multi-agent setup  like in~\cite{Waschneck.2018} and~\cite{Park.2020}. The transition from single- to multi-agent RL is also non-trivial in general as thoroughly discussed in~\cite{Kapoor2018}. Specifically, as the individual agents change their behavior during learning, the environment of each agent becomes non-markovian resulting in a partially observable Markov decision process (POMDP). Early approaches use cooperative settings based on Q-learning assuming that the actions of other agents are made to improve the collective reward~\cite{Lauer2000,Matignon2007}. In recent years, various approaches based on deep MARL have been introduced~\cite{Lowe2017,Foerster2017, Sunehag2017,Omidshafiei2017}. 

All the above-mentioned RL advancements either use convolutional neural networks~\cite{Krizhevsky.2012} or a Multi-Layer Perceptron (MLP), which uses feature engineering to achieve performance. Also, deep learning methods do not fare well on data in the non-Euclidean space, and CNN and MLP are inherently neither size nor permutation invariant. The addition of a single machine or job to the scheduling environment requires training the agent completely from scratch since the previously trained notions of state are not valid anymore. GNNs are size and permutation invariant and additionally, they also provide information on pair-wise interaction between the objects~\cite{Battaglia.2018}. ~\cite{JoanBruna.2014} developed an input size invariant network for image classification, while~\cite{Zaheer.2017} developed an algorithm that is generally applicable to estimation, classification, and outlier detection problems. GNNs on reinforcement learning in scheduling problems have also been developed~\cite{Seito.2020}~\cite{ParkJunyoung.2020}. While~\cite{Seito.2020} focuses on the computational performance,~\cite{ParkJunyoung.2020} focuses on the performance but their results are only compared with priority dispatch rules (PDRs), one of them being FIFO.

GraSP-RL focuses on the generalization of the trained agents. By generalization, we mean the resilience of the agent trained using GraSP-RL to perturbations in the available jobs and machines. We accomplish this by applying changes in the JSSP environment to obtain two new classes of problems: an OSS and an RJSSP. In OSS, additional machine capacity is added to the production unit. So now the job can be processed on two similar machines instead of just one in the JSSP. Next, we also created five scenarios for RJSSP where jobs are removed randomly from the plan and the agent is tested without any further training. The plans made by the RL agent in all the environments are then compared with algorithms like TS and GA. The results show that the RL agent performs better than the TS and GA in almost all scenarios. To the best of our knowledge, ours is the first paper to compare the performance of the RL agent to the metaheuristics like TS and GA.

\section{Preliminaries}\label{sec:Prelims}

\subsection{Job Shop Scheduling Problem}

The goal of a scheduling algorithm is to use a limited number of resources or machines $m \: \in \: \{m_1, m_2, m_3, ..., m_m\}$ with each machine having a unique processing capability, to process a set of jobs $j\: \in \: \{j_1, j_2, j_3, ...,j_n \}$, while optimizing a specified objective like make-span, tardiness reduction, due date violation reduction, etc. In this paper, we focus on the objective make-span reduction $C_{max}$. Each of these $j$ jobs have $k$ operating sequence $o\: \in \: \{o_1^j, o_2^j, o_3^j, ..., o_k^j\}$, through the machines with a specified processing time $t_{m, \:j, \:k}$ at the corresponding machine. When the final operation is completed the job is considered finished. The preceding operation has to be completed for the next operation to start and the machines can only process one job at a time.

\subsection{Reinforcement Learning}

\begin{figure}
    \centering
    \includegraphics[scale=0.5]{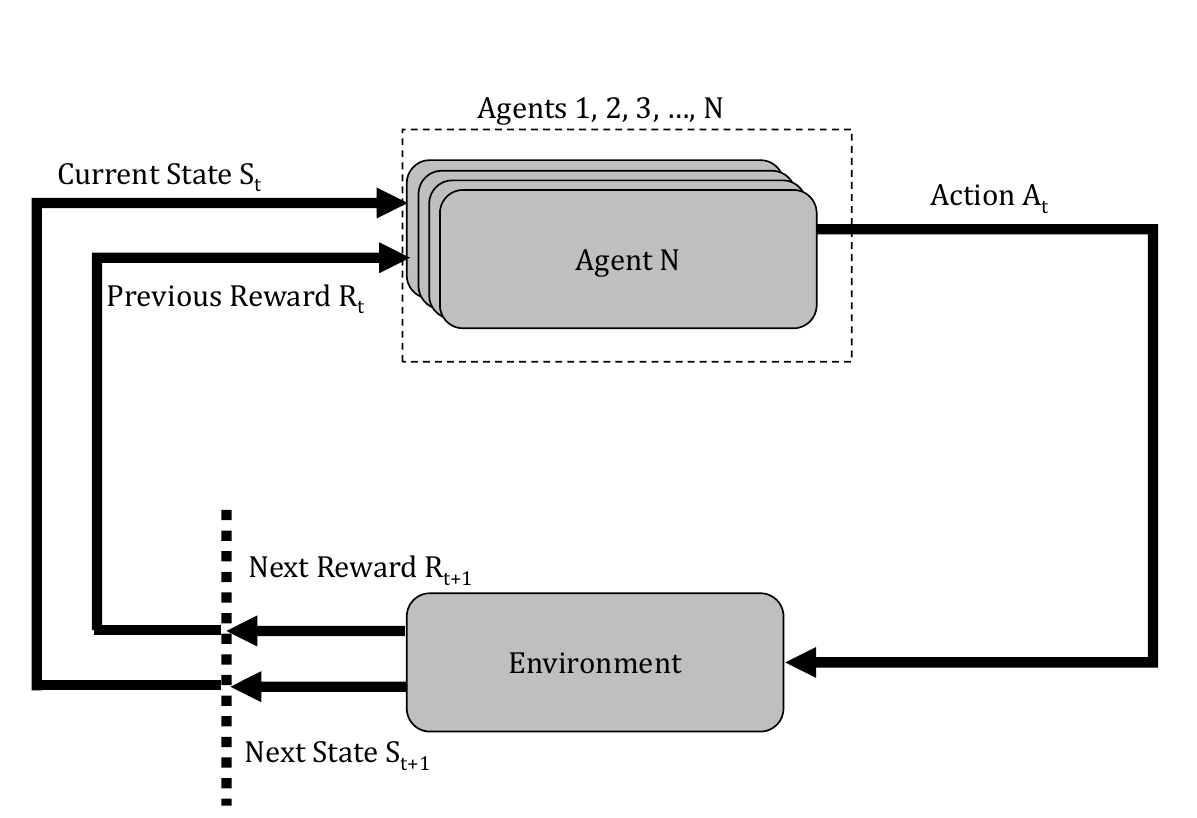}
    \caption{The interaction of agent and environment~\cite{Sutton.2018}.}
    \label{fig:rl}
\end{figure}

Reinforcement Learning (RL) is a branch of machine learning dealing with training agents in an environment, where the agent is not explicitly told what to learn but rather has to learn through its interaction with the environment. The agent provides an action $a \: \in \: \mathcal{A}$ as a response to a state $s \: \in \: \mathcal{S}$ in the environment. The state of the environment changes as a response to this action and the agent receives the new state $s' \: \in \: \mathcal{S}$ and a reward $r$ for the action previously taken. This is pictorially shown in Fig.~\ref{fig:rl}. The agent's objective is to maximize the long-term collection of this reward.~\cite{Sutton.2018} describes RL as, \say{learning what to do – how to map situations to actions – so as to maximize a numerical reward signal}.  The problem of RL is formalized as a Markov decision process (MDP) and the dynamics of the MDP can be defined by the tuple $\mathcal{(S, A, P, R,})$, where $\mathcal{S}$ is the set of states, $\mathcal{A}$ is the set of actions, $\mathcal{P}$ is the state transition model for a given action, and a reward function defined by $\mathcal{R: S \times A \times S \xrightarrow{} \mathbb{R}}$ for each state transition $s_t \xrightarrow{} s_{t+1}$ because of the action $a_t$.

Estimating the value functions is one of the fundamental things in RL. Value functions provide an idea of \textit{goodness} of being in a state $v(s) = \mathbb{E}[R_{t}|S_{t}=s]$, or executing a state-action pair $q(s, a)  = \mathbb{E}[R_{t}|S_{t}=s, A_{t} = a]$. This is based on the rewards that the agent can expect to earn, now and in the future, following a certain policy $\pi(a|s)$. The policy is the agent's mapping of the states in the environment, $s$, to the probabilities of the agent's actions, $a$. The value function $R_t$ is given by $\sum_{t}{\gamma^t r_{t}}$, where $\gamma \: \in \: [0,1]$ is the discount factor that weights the importance of rewards in future and $r_t$ the step-wise reward as collected by the agent as time $t$.
  
\begin{equation}
    \label{eq:policy_update}
    \theta_{t+1} = \theta_{t} + \rho \nabla J(\theta_{t}),
\end{equation}

RL methods can be split into two: tabular methods and approximate solution methods. Although tabular methods are powerful enough for simple problems, they are not scalable for problems with bigger state space and action space. To overcome this issue, approximate solution methods introduce a function approximator, such as neural networks, to overcome the scalability issues of tabular methods. Here, the agent approximates the policy directly without the need for a value function. Approximating both the value function and the policy, although, has been shown to improve efficiency drastically, like in Advantage Actor-Critic (A2C)~\cite{VolodymyrMnih.2016}.

The policy maximizes the rewards by  optimizing a performance function $J(\theta_{t})$ and the trainable parameters $\theta$ of the policy are updated using Equation~\eqref{eq:policy_update}. In this paper, we use an approximate solution algorithm called the Proximal Policy Optimization (PPO)~\cite{Schulman.2017}. PPO builds on the work of~\cite{schulman2015}, where the magnitude of the updates to the agent is controlled by using trust region constraints. PPO uses a computationally simple approximation to achieve the same.

\begin{equation}
    \label{eq:ppo}
    \underset{\theta}{max} \: \mathbb{E}_{\sim \pi_{\theta}} \left[ min ( r_{t}(\theta)\hat A_{t}, \: clip(r_{t}(\theta), 1-\epsilon, 1+\epsilon) \hat A_{t})  \right]
\end{equation}
where
\begin{equation}
    r_{t}(\theta) = \frac{\pi_{\theta}(a_{t}|s_{t})}{\pi_{\theta_{old}}(a_{t}|s_{t})}
\end{equation}

Equation~\eqref{eq:ppo} shows the objective function optimized in PPO, where $r(\theta)$ is the probability ratio, $\hat{A_t}$ is the advantage function, and $\epsilon$ is the clipping hyperparameter. There are two terms in the objective function. The first term is the function we ideally want to optimize. Intuitively, we want to increase the probability of the actions that have a high advantage value. But optimizing this function alone causes stability issues in the learning process. Hence, a second term is introduced that bounds the ratio by a factor of $1\pm\epsilon$, i.e., if the ratio is higher than $1+\epsilon$ or lower than $1-\epsilon$, then the ratio is clipped. And by taking a minimum of the first and the second term, the final objective function is a lower bound on the first term. The objective function of PPO ensures small and incremental updates to the policy, thus stabilizing the learning process.

\subsection{Gradient Monitoring}
While in supervised learning the targets do not change and the labels/classes for a data point remain the same throughout the training, in RL the cost function changes dynamically as the agent explores the unseen parts of the environment. This makes RL a naturally noisy learning process. The variance in the gradients between updates is high and can sometimes lead to destabilization of the learning process. Further, the non-isomorphic graphs generated by the production environment lead to high variance in the feature extracted by the GNNs. Such high variance in the input and the learning process causes instability in the training of the RL agent.

To overcome these problems, a regularizer called Gradient Monitoring (GM) \cite{Hameed2021} is utilized. GM applies a further constraint on the trust region of the gradient updates by significantly reducing the gradient variance. GM has been shown to improve performance in standard RL environment settings such as Atari Learning Environment (ALE)~\cite{Bellemare2013} and MuJoCo~\cite{Todorov2012}. GM uses a Heaviside step function on the ratio between the weights and their corresponding accumulated gradients to decide if the gradients are important enough to pass through. This is shown in Equation~\eqref{eq:heaviside} and \ref{eq:gm_momentum}.

\begin{equation}\label{eq:heaviside}
	M_{t} =  {H} (D_{t} - \lambda \mu),
\end{equation}

\begin{equation}\label{eq:gm_momentum}
	M_{\zeta} =  M_{\zeta}\zeta + M_{t}(1-\zeta) 
\end{equation}

where $M_{t}$ is the decision matrix, $\mu$ is the  mean of $(\left|\frac{\nabla L_{W_t}}{W_t}\right|)$, $\lambda$ is the the threshold hyperparameter, $M_{\zeta}$ is the momentum matrix, and $\zeta$ is the momentum value. The decision matrix $M_{t}$ decides if the gradient signal is strong enough to update the weight. The momentum matrix $M_{\zeta}$ then updates the momentum for each gradient value. The application of momentum in GM reduces the gradient norm in the updates, thereby increasing the agent's learning  performance. Another advantage is the reduction in the influence of the network size of the agent since the threshold hyperparameter $\lambda$ in GM effectively controls the number of active learning parameters during training.

\subsection{State-space encoding for RL}

RL agents learn by interacting with an environment (the JSSP environment in this paper). The environment provides the state while the agent uses that state information to take an action on the environment. The JSSP environment can represent the state as a feature vector that describes the production system using encoding techniques~\cite{Gabel.2008}. It is crucial that the state representation captures the information required in the current setup and the general problem class.

While encoding the global JSSP information provides a reasonable representation of the system, it fails to capture the relationship between the entities of JSSP, like machines and jobs. It could be argued that this can be learned within the MLP, but MLPs provide a weak implicit relational inductive bias after training. Hence we develop GraSP-RL where the production environment information is converted into a graph and then it is further processed by the GNNs, to extract an encoding that has a strong inductive bias~\cite{Battaglia.2018}.

\section{Introduction to GraSP-RL}\label{sec:jsspgraph}

\subsection{Modeling JSSP as Graph}

\begin{figure}[h]
    \centering
    \includegraphics[width=\linewidth]{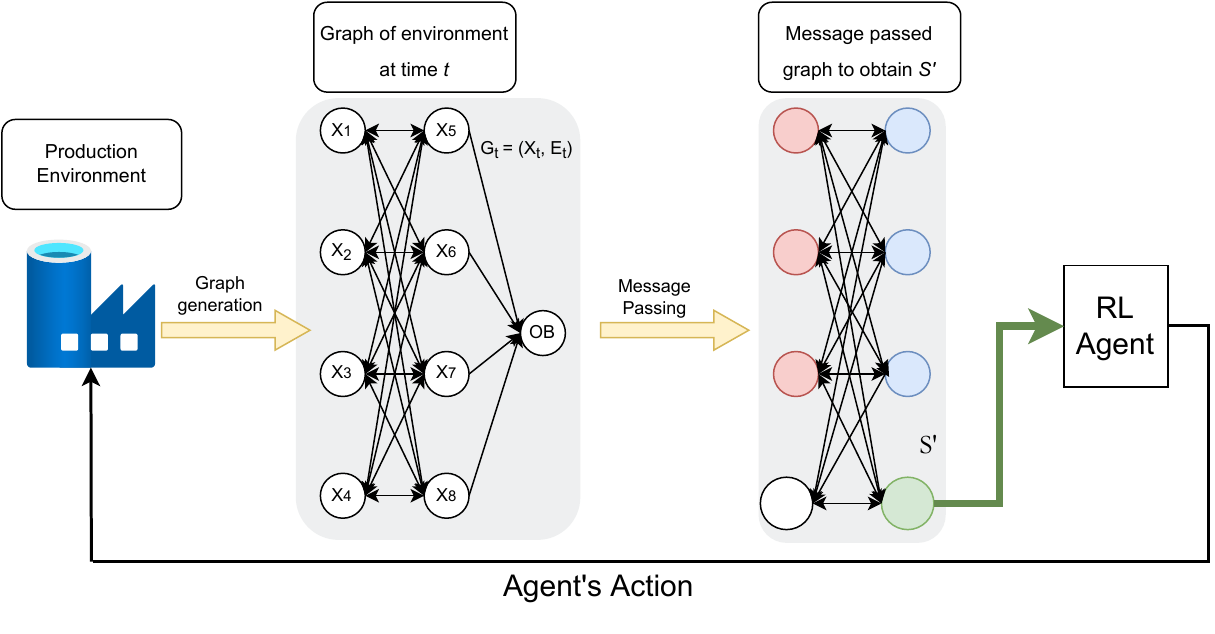}
    \caption{Workflow of GraSP-RL represented pictorially. }
    \label{fig:framework}
\end{figure}

In this section, we introduce GraSP-RL which is pictorially shown in Figure~\ref{fig:framework}. The cycle starts when a machine is free and a job is to be loaded into it from its buffer. The current data from the production environment is first used to construct a graph of the environment state, with their corresponding node and edge information. The constructed graph is then sent through a GNN. The neighborhood information is aggregated in all the nodes. After this step, the relationship between the machines and the remaining jobs is captured in each node from its point of view. The feature vector of the buffer of the current empty machine (X8 in Figure~\ref{fig:framework}) is then passed on to the RL agent (PPO) to take an action. The action tells the environment which job from the corresponding machine buffer is to be loaded into the machine. The cycle repeats when the next machine becomes free.

On the environment side, a progressive curriculum learning strategy is used to feed the agents with simpler scenarios first. And then increasingly difficult scenarios are generated, ultimately ending with the final scenario of 30 jobs. The agent is used in a distributed manner, where each machine is assigned an agent. But they are updated centrally and then shared again, which allows the agent to encounter a diverse set of scenarios. A detailed explanation of the various parts of GraSP-RL is given below.

\subsection{JSSP as graphs}

Graphs are a natural way to represent JSSP. They capture the relationship between the machines, machine buffers, and the jobs. A graph is defined as a tuple $G = (X, E)$, where $X$ is the vertices or nodes and $E$ is the directed edges connecting the nodes. We define $X \supseteq \{ X_{mb}, X_{m}\}$, where $X_{mb} \subset \{X_1, X_2, X_3, ..., X_m \}$ is the set of machine buffers, and $X_m \subset \{X_{m+1}, X_{m+2}, X_{m+3}, ..., X_{2m}\}$ is their corresponding set of machines. $E = (e_{sr},\: s_j,\: r_j) \in j$, where $e_{sr}$ is the edge feature, and $(s_j,\:r_j) \in X$. The cardinality of $X$ and $E$ are given by $N^x$ and $N^e$ respectively. The edges are calculated dynamically at each time step based on the jobs available in the environment and their remaining operation sequence.

\begin{figure}[htp]
    \centering
    \includegraphics[scale=0.5, align=c]{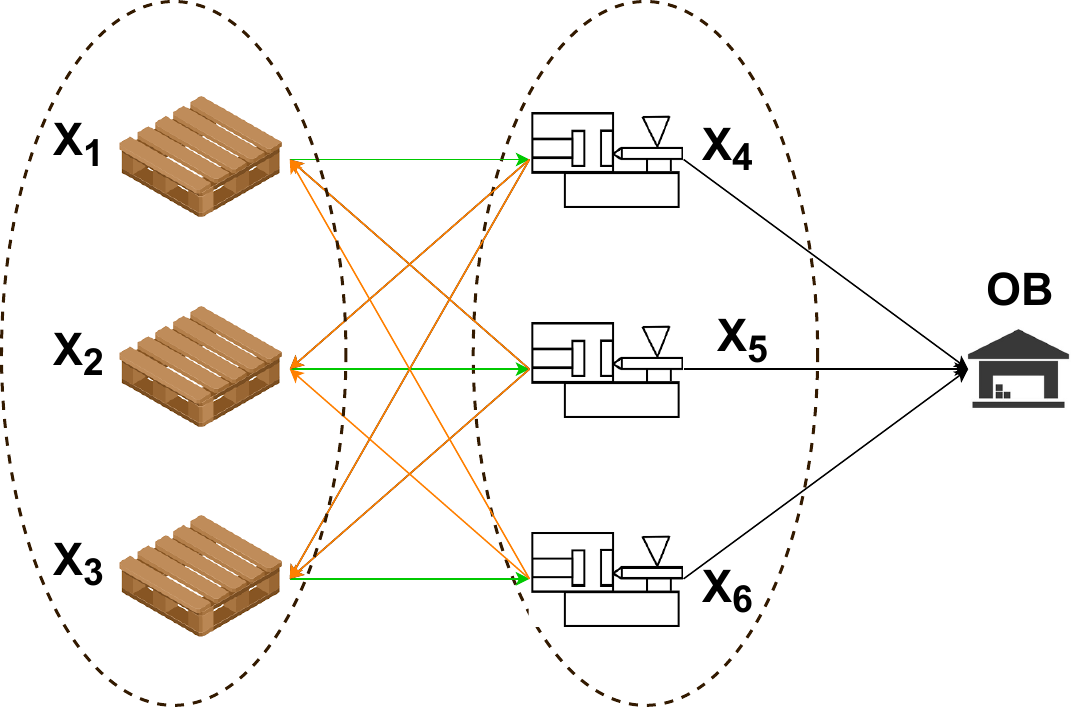}
    \caption{The general representation of a JSSP as a bipartite graph is shown. Every machine node ($X_4$ to $X_6$) has its own buffer node ($X_1$ to $X_3$). All the possible directed edges connecting the nodes are represented with a pointed arrow. Orange arrows represent the outward job movement while green represents the inward job movement. $OB$ is the output buffer, where all the finished jobs are stored. Jobs present in the machine node and machine buffer node are represented as a node feature vector.}
    \label{fig:IMM_setup}
\end{figure}

\textbf{Bipartite Graph:} In the graph generated by GraSP-RL, the machine nodes are always preceded by their corresponding machine buffers in the operation sequence of any job. And since the set of all vertices X(G) is a union of two disjoint independent sets $X_m$ and $X_{mb}$, as per the definition the generated graph G is bipartite. An example of the graph is shown in Figure~\ref{fig:IMM_setup}. The bipartite nature of the generated graph permits message passing between only a machine and its connected machine buffers or vice-versa.

\textbf{Node features and dynamic edges:} Every node in the graph generated by GraSP-RL has a node feature associated with it. The machine nodes $X_m$ are a one-hot encoded vector of the job it is currently processing, as shown in Equation~\eqref{eq:node_feature_machine}.

\begin{equation}
    e_i(x) = 1_A (x) :=
    \begin{cases}
    1, & \text{if}\ x \in A\\
    0, & \text{if}\ x \notin A
    \end{cases}
\end{equation}

\begin{equation}
\label{eq:node_feature_machine}
X_m  = e_i(j_m) \text{, where $j_m$ is the current job.}
\end{equation}

While for the machine buffer nodes $X_{mb}$ it is the summation of all the one-hot encoded vectors of the jobs that it holds, shown below.

\begin{equation}
\label{eq:node_feature_buffer}
X_{mb} = \sum_{n=0}^{j_{mb}} e_i(n)\text{, where $j_{mb}$ is the set of jobs.}
\end{equation}

The node features are updated according to the job present in the machine or the machine buffer, while the edges are created dynamically based on the job present in a machine. For example in Figure~\ref{fig:IMM_setup}, if a job present in $X_1$ has operations in machines $X_4 \text{ and } X_5$ sequentially, then it generates the edges $\{e_{X_1 X_4}, e_{X_4 X_2}, e_{X_2 X_5}\}$. The node features $X$ and the edges $E$ generated then represent the current state of the JSSP as a graph $G = (X,\:E)$. The graph $G$ is then passed into a GNN to extract the encoded features for that machine.

GraSP-RL generates a non-isomorphic graph at each step of the environment owing to the nature of the node features, as proved in Theorem~\ref{th:1}, and dynamic edge generation. The generated graph when processed through GNN embeds any two non-isomorphic graph states of the JSSP environment into different feature embeddings. The detailed working of GNN is explained in Section~\ref{subsec:GNN_proc}

\begin{definition}[Isomorphic Graphs]
\label{def:isomorph}
    An isomorphism from a simple graph G to a simple graph H is a bijection $f: X(G) \rightarrow X(H)$ such that $u v \in E(G)$, the edge set of graph G, if and only if $f(u) f(v) \in E(H)$~\cite{douglaswest1996}.
\end{definition}

\begin{corollary}[Non-isomorphic Graphs]
    Non-isomorphic graphs are topologically dissimilar to each other, i.e. no bijection can map the two graphs.
\end{corollary}

\begin{theorem}
\label{th:1}
Given a unique set of node features, and an injective function in the message-passing algorithm of GNN, the GNN generates unique node embeddings.
\end{theorem}

\begin{proof}
It has been established that GNNs are as powerful as the Weisfeiler-Lehman graph isomorphism test (WL test) if the neighborhood aggregation function is injective~\cite{xu2018powerful}. Thus if the node feature set input to the GNN is unique, then GNN generates unique node embeddings. Remember, the set of all node features is uniquely one-hot encoded as shown in Equation~\ref{eq:node_feature_machine} and~\ref{eq:node_feature_buffer}. So it can be proved by contradiction that any subset of the unique input node feature set is unique. 

Consider a set of unique nodes $X$. Now consider that an element $y \in X$ is not distinct in the set of all subsets $\mathcal{P}\{X\}$. Since $X$ is the set of distinct nodes, the element that makes $y$ indistinct must also be present in $X$, contradicting the definition of $X$ itself.

This proves that $y \not \subseteq X$ and further, the assumption that $y$ is indistinct is false. Thus the subsets of $X$ are always distinct. The input of a distinct subset to an injective function will always generate unique embeddings, owing to the definition of injective functions which map distinct inputs to distinct outputs.
\end{proof}

\subsection{GNN processing of the graph}
\label{subsec:GNN_proc}
The strength of GNNs is in the discriminatory power of their encoding to differentiate between two non-isomorphic graphs. GNNs can be powerful discriminators of non-isomorphic graphs by using even simple GNN architecture under certain conditions~\cite{xu2018powerful}. Deep learning revolutionized many machine learning tasks like object detection, image classification, image segmentation, machine translation, etc. which are all represented in Euclidean space. But graph-structured data are typically present in the non-Euclidean space. Hence, they require the special architecture of GNNs. The existing neural architectures do not effectively use the inductive bias present in graph-structured data. GNNs overcome this by establishing a strong relational inductive bias by exploiting the graph-structured data rather than being hindered by it~\cite{Battaglia.2018}.

\textbf{Message Passing Algorithm:} GNNs use an algorithm called message passing at a particular node that iteratively aggregates neighborhood node information and updates the node representation. This is done in three stages, AGGREGATE, COMBINE, READOUT. In AGGREGATE, the neighboring nodes $\mathpzc{N}(i)$ of the node $X_i$ are passed through a function and summed over all neighboring nodes, as shown below,

\begin{equation}
    \label{eq:aggregate}
        a_i^k = \sum_{j=0}^n ACT(X_j), \: \forall \: j \in \mathpzc{N}(i)
\end{equation}

 We use rectified linear (ReL) activation function in $ACT$. The other choices include $Tanh, Sigmoid, PReLU, etc.,$ The aggregated value $a^k_i$ is then passed onto the COMBINE function, where is it summed with the node feature vector from the previous time step $X_i^k$, as shown below,

\begin{equation}
    \label{eq:combine}
        X^k_i = \big( a_x^k + X^{k-1}_i\big)
\end{equation}

This process is repeated for k-iterations after which the node vector captures the node information within its $k$-hop neighborhood and which is then passed into the READOUT function, where the node/machine on which the action needs to be taken is passed through a differentiable function $\psi$, as shown below, 

\begin{equation}
    \label{eq:readout}
        S' = \psi \: \big( X^k_i\big)
\end{equation}

\subsection{Multi-agent reinforcement learning}

The node feature is now encoded in a lower dimensional Euclidean space compared to the non-Euclidean space of the graph structure. There are two advantages of using graph representation for JSSP and processing the graph using GNNs:

First, it reduces a global optimization problem to a local optimization one, i.e., at any point in the interaction between the agent and the environment, we are only focused on the current free machine, as already shown in Fig.~\ref{fig:framework}. $S'$ enables the RL agent to act on the environment in a distributed manner and is end-to-end trainable. By distributed, we mean the same agent taking action on multiple machines. The generalization performance of the GraSP-RL in OSS and RJSSP can be attributed to the powerful neighborhood aggregation done by the GNNs. This ensures that even if machines are added the node feature that the agent uses contains the information of just its neighbors, thus making it a local optimization problem.

Second, the state space only scales linearly with an increase in the number of jobs, and its dimension is detached from the number of machines, unlike the previous approaches. Hence our approach enables the neural network to be smaller in size, leading to better sample efficiency~\cite{Ha2018}.

\subsection{Curriculum Learning}

Neural networks are at the core of modern machine learning tasks, including RL, which are analogous to the neurons in the human brain. So it only seems logical that we train the RL agent like we would teach a human. Instead of initiating training with tasks of random difficulty levels, curriculum learning sorts the tasks based on their difficulty level, and the simplest tasks are first shown to the agent and increasing the difficulty as it continues training. This gradual increase in complexity enables the agent to find a good local minimum wherein it generalizes well for future complex tasks as well. Curriculum Learning has been shown to improve generalization performance~\cite{bengio2009}.

\begin{figure}
    \centering
    \includegraphics{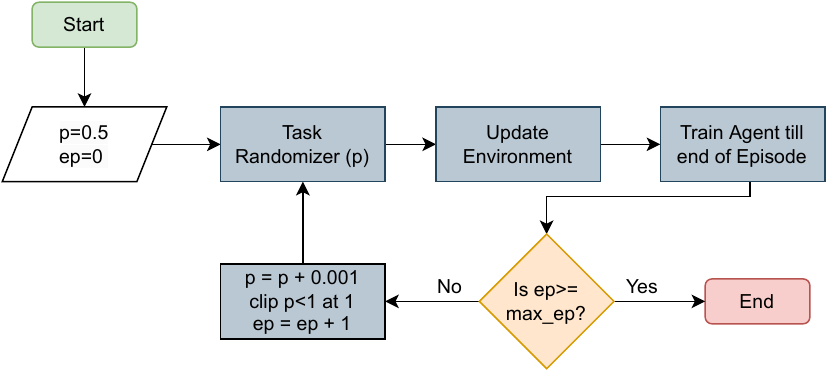}
    \caption{Flow depicting the flowchart for the Curriculum Learning process.}
    \label{fig:cl_flowchart}
\end{figure}

The curriculum learning applied here focuses on the availability of jobs. This is pictorially shown in Fig.~\ref{fig:cl_flowchart}. At the start of the training, all jobs are sampled with a probability of p=50\%, which is done through a \textbf{task randomizer}. The idea is that with fewer jobs, it is easier for the agent to make good plans while also learning about the inter-machine relationships that affect the setup time in the jobs. The randomly picked jobs are then used to initiate the environment in which the agent trains. As the agent continues training the probability of a job's availability increases at a rate of 0.1\%.

The probability p is capped at 100\%, at which point the agent begins making the actual plan with all the jobs available. The agent continues training with the full availability of jobs before the training is terminated at the preset number of episodes. The curriculum learning is active for 39\% of the training after which all the jobs are available.

The next section discusses the implementation of GraSP-RL along with a detailed explanation of the environment in which it is implemented. The coding is completely done in python and we use open-source packages for the same. For RL, we use a python library called Stable Baselines3~\cite{stable-baselines3} which internally uses Pytorch~\cite{Paszke2019}, and for GNN, PyG~\cite{Fey2019} is used.

\section{Experiments and Results}\label{sec:expres}

\subsection{Injection molding machine environment}

The above-described framework, GraSP-RL, is then applied in an environment consisting of 30 jobs, 4 machines, 4 machine buffers, and 1 output buffer. While passing the graph information into the GNN, the node features of the output buffer are not used. Each machine has a machine buffer in front of it. Each of the 30 jobs has a unique sequence of operations through the machines. At initialization, the machine buffer nodes are populated with the jobs whose first operation is in their corresponding machines. Table~\ref{tab:machine_buffers} shows the jobs loaded in each of the machine buffers. Every job and machine combination has its own processing time with a setup time included if there is a setup change.

\begin{table}[htp]
\centering
\begin{tabular}{cccc}
\hline
Machine Buffer-1 & Machine Buffer-2 & Machine Buffer-3 & Machine Buffer-4 \\ \hline
7    & 4    & 1    & 3    \\
11   & 8    & 2    & 5    \\
15   & 14   & 6    & 10   \\
18   & 23   & 9    & 12   \\
22   & 25   & 13   & 16   \\
29   & 28   & 17   & 19   \\
     &      & 21   & 20   \\
     &      & 24   & 26   \\
     &      & 27   &      \\
     &      & 30   &      \\ \hline
\end{tabular}
\caption{List of jobs that are populated in each machine buffer at initialization.}
\label{tab:machine_buffers}
\end{table}

The jobs are split into three types: type-1 (1-10), type-2 (11-20), and type-3 (21-30). They are indicated by the shades of red, green, and blue respectively in the Gantt charts shown below. The machine needs a setup change, and therefore a setup time, if the previous job and current job are from two different types. When the environment is initialized, the jobs are loaded into the machines where their first operation is to be performed. From then on, they are moved through the environment based on their operation sequence. The objective of the RL agent interacting with the environment is to make a plan with the least possible makespan.

When the remaining processing time of a job in a machine becomes zero, the job is moved out of the machine, and the machine is designated as empty and ready for the next job. The movement of the job from one machine to another machine buffer after the completion of the processing is done automatically since no decision is to be made here. At this point, the graph representation of the JSSP environment is generated with its corresponding node features and dynamic edge index. The generated graph is then passed to the GNN where message passing is applied and the feature vector of the empty machine is then further passed to the RL agent. The agent's decision-making is applied to the movement of the jobs from the machine buffer to its corresponding machine. After the job, which the agent selected, is loaded on the machine, the cycle continues.

The environment resets to the initial values if it reaches either a positive or a negative terminal state. A positive terminal state is when the state indicates the successful completion of the environment's objectives (i.e. if all 30 jobs are moved into the output buffer after the completion of all their respective operations). A negative terminal state is when the agent is unable to plan all the jobs within a predefined maximum number of steps.

\subsection{Results for JSSP in the injection molding machine environment}

\begin{figure}[h]
    \centering
    \includegraphics[width=\linewidth]{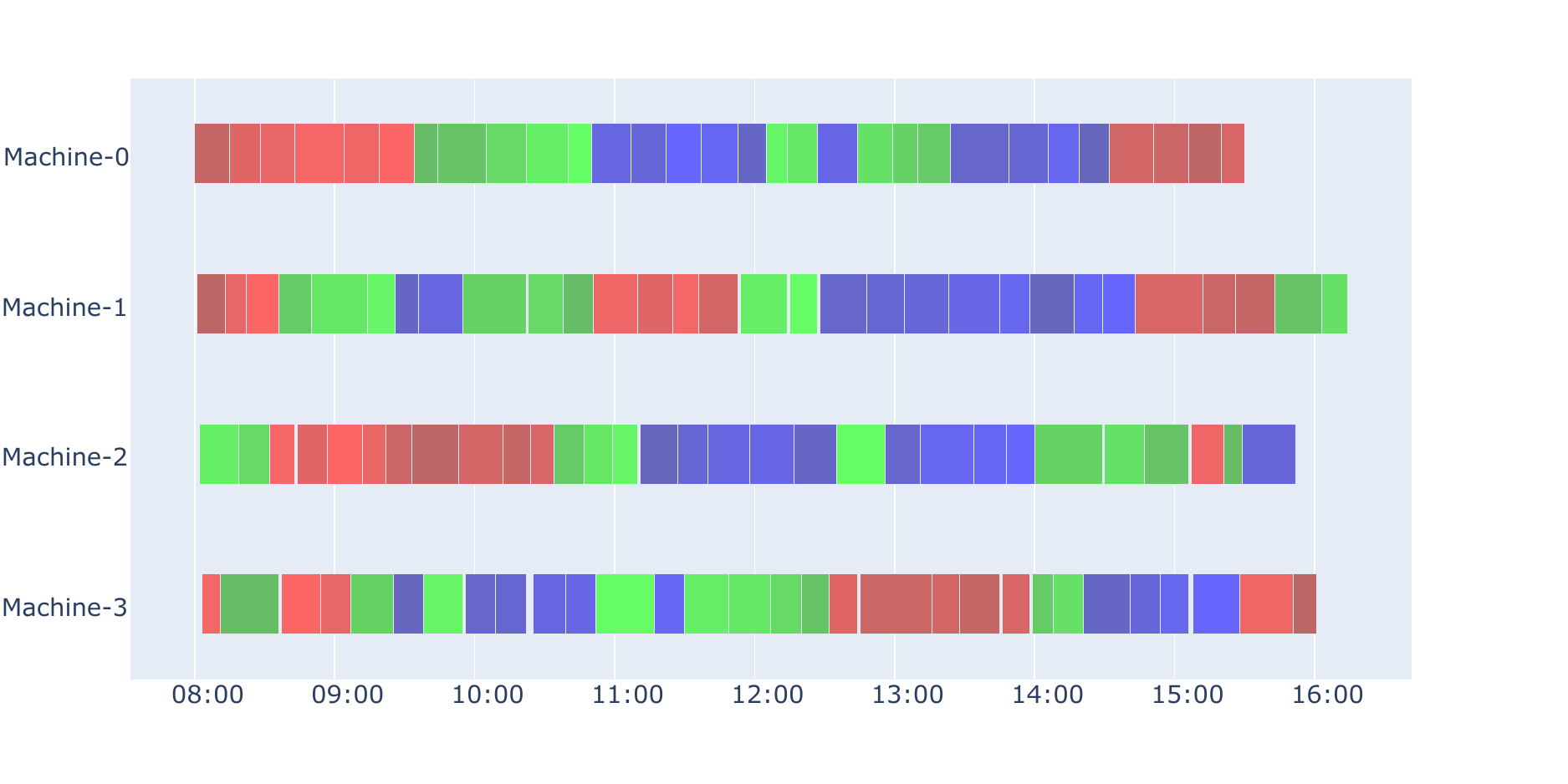}
    \caption{Gantt chart of the plan made by GraSP-RL.}
    \label{fig:GNN_JSSP_GANTT}
\end{figure}

The RL agent's training and testing performance depend on a number of hyperparameters of the PPO algorithm. To optimize the agent's performance, the python library Optuna~\cite{optuna2019} is used. The optimized hyperparameters of PPO are then used to train and save an RL agent. The plan made by the RL agent is shown in Fig.~\ref{fig:GNN_JSSP_GANTT} with a makespan of 494 minutes.

\subsection{Comparison with Tabu Search and Genetic Algorithm}

\begin{figure}[htp]
    \centering
    \includegraphics[width=\linewidth]{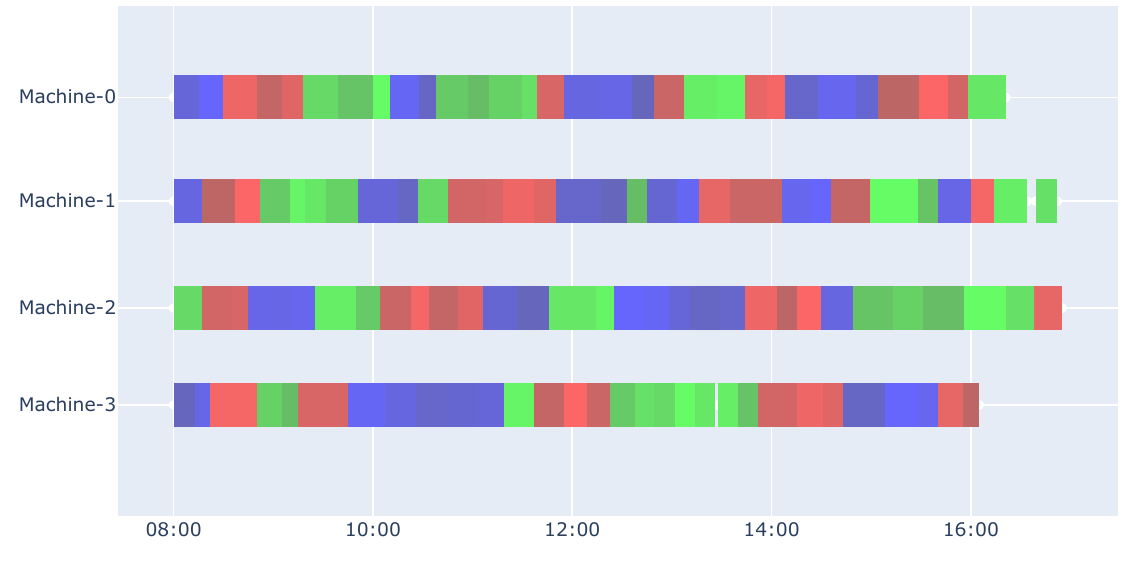}
    \caption{Gantt chart of the plan made by the tabu search.}
    \label{fig:TS_JSSP_GANTT}
\end{figure}

\begin{figure}[htp]
    \centering
    \includegraphics[width=\linewidth]{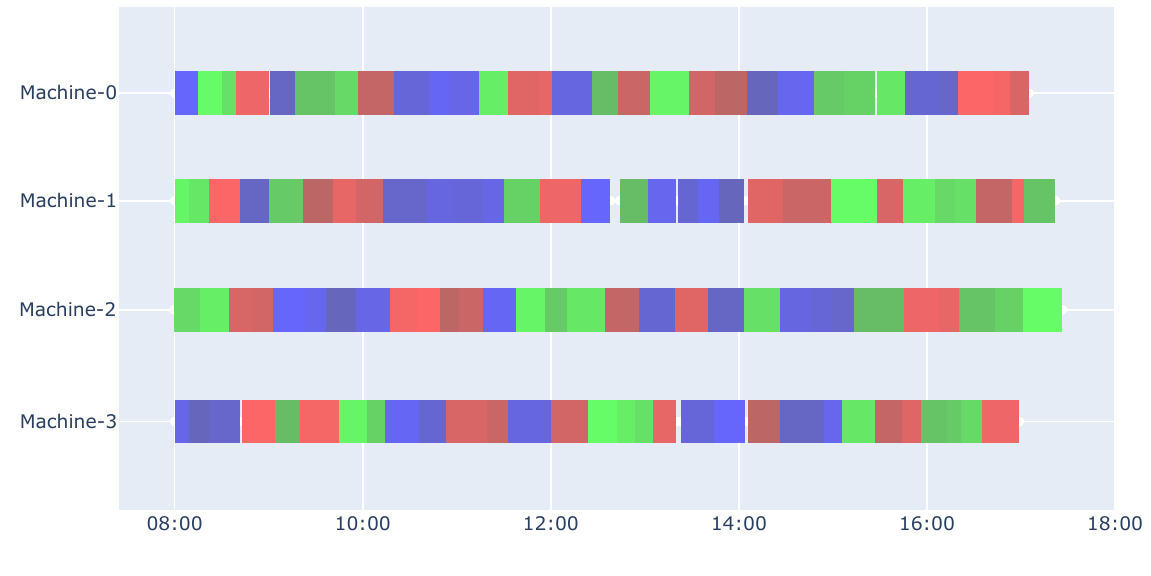}
    \caption{Gantt chart of the plan made by the genetic algorithm.}
    \label{fig:GA_JSSP_GANTT}
\end{figure}

To compare the RL agent's performance with the TS and GA, we use an open-source python library called JSSP~\cite{mcfadd2020}. The best plan made by the TS and GA is shown in Fig.~\ref{fig:TS_JSSP_GANTT} and~\ref{fig:GA_JSSP_GANTT} respectively. Since the plan made by TS and GA is stochastic in nature between runs, we optimized the parameters and ran 100 trials for each of the methodologies. The average makespan for TS and GA were 518.7$\pm$6.5 minutes and 566.5$\pm$10.6 minutes respectively. Further to establish a baseline result, the JSSP environment was also solved with FIFO (611.7$\pm$21.9 minutes) and random actions (591$\pm$18.8 minutes). In FIFO, the order execution is prioritized and the job with the highest priority is always loaded into the machine, while in random actions, the actions are purely random.

The plan made by the RL agent is superior to the rest of the methods while also being deterministic. It can be seen that while in Figure~\ref{fig:GNN_JSSP_GANTT} only one machine has processing times more than 16:00, in Figure~\ref{fig:GA_JSSP_GANTT} and \ref{fig:TS_JSSP_GANTT} all the machines go beyond the 16:00 time. This could be attributed to the increased setup times in the plan made by TS and GA. While GraSP-RL manages to complete the plan with only $37$ setup changes, TS and GA use $48$ and $64$ setup changes respectively, as shown in Table~\ref{tab:setup_comparison}. The performance of GraSP-RL can be attributed to it sacrificing $Machine-4$, where there are more jobs with setup time, to accommodate fewer setup times in other machines, which is possible because the agent is informed of what happens in the other machines through the message-passing of node information. The comparative performance improvement by GraSP-RL is shown in Fig.~\ref{fig:all_jobs_ms_compare}, where the baseline is the RL agent's makespan.

\begin{table}[htp]
\begin{tabular}{cccc}
\hline
                     & GraSP-RL & Tabu Search & Genetic Algorithm \\ \hline
Make span            & 494 $\pm$ 0.0 & 518 $\pm$ 6.5 & 566 $\pm$10.6 \\
No. of setup changes & 37            & 48          & 64              \\ \hline
\end{tabular}
\caption{Comparison of the setup changes done by TS, GA, and GNN\_RL algorithms.}
\label{tab:setup_comparison}
\end{table}

\subsection{Ablation study for GraSP-RL}

\begin{figure}[htp]
    \centering
    \includegraphics[width=\linewidth]{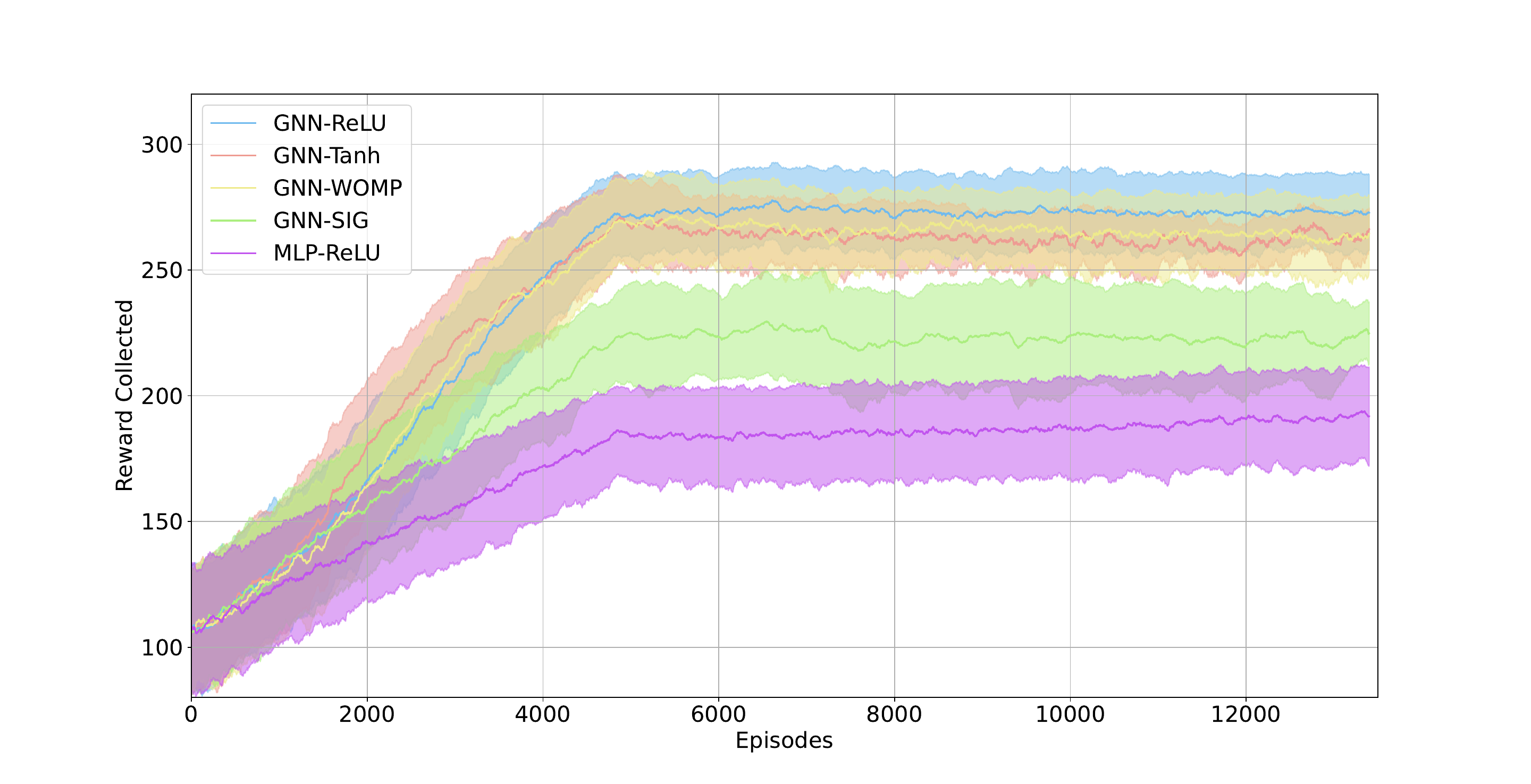}
    \caption{Rewards collected by each method in the ablation study.}
    \label{fig:ablation_study}
\end{figure}

To also empirically prove the effectiveness of GraSP-RL against similar graph and non-graph methods, we conduct an ablation study. The training performance of GraSP-RL is compared with three other methods. First, named MLP-ReLU, without the GNN using only the one-hot encoded state information passed directly into an MLP. The second, named GNN-WOMP, uses the one-hot encoded current machine information but without the GNN. The third, named GNN-ReLU, GNN-Tanh, and GNN-SIG, use a variety of activation functions like \textit{Sigmoid, Tanh, and ReL}.

As shown in Figure~\ref{fig:ablation_study}, GraSP-RL outperforms the other methods with a p-value of close to zero in Welch's t-test. Here, the rewards collected are inversely proportional to the makespan $C_{max}$. It is to be noted that in the first 5000 episodes, curriculum learning is in effect, after which the agents train for an additional 8000 episodes without any significant improvement in performance. It can also be seen that all the graph methods outperform the non-graph method MLP significantly, underlining the importance of the graph structure of the production data. 

\subsection{Testing for generalization performance on OSS and RJSSP}

To test the generalization capability of the RL agent, we also test the RL agent on scenarios it was not exposed to. By test, we mean deploying the agent in those scenarios without any further training and directly extracting the plan. This generalization capability is one of the key contributions of this paper.

\begin{figure}[htp]
    \centering 
\begin{subfigure}{0.5\textwidth}
  \includegraphics[width=\linewidth]{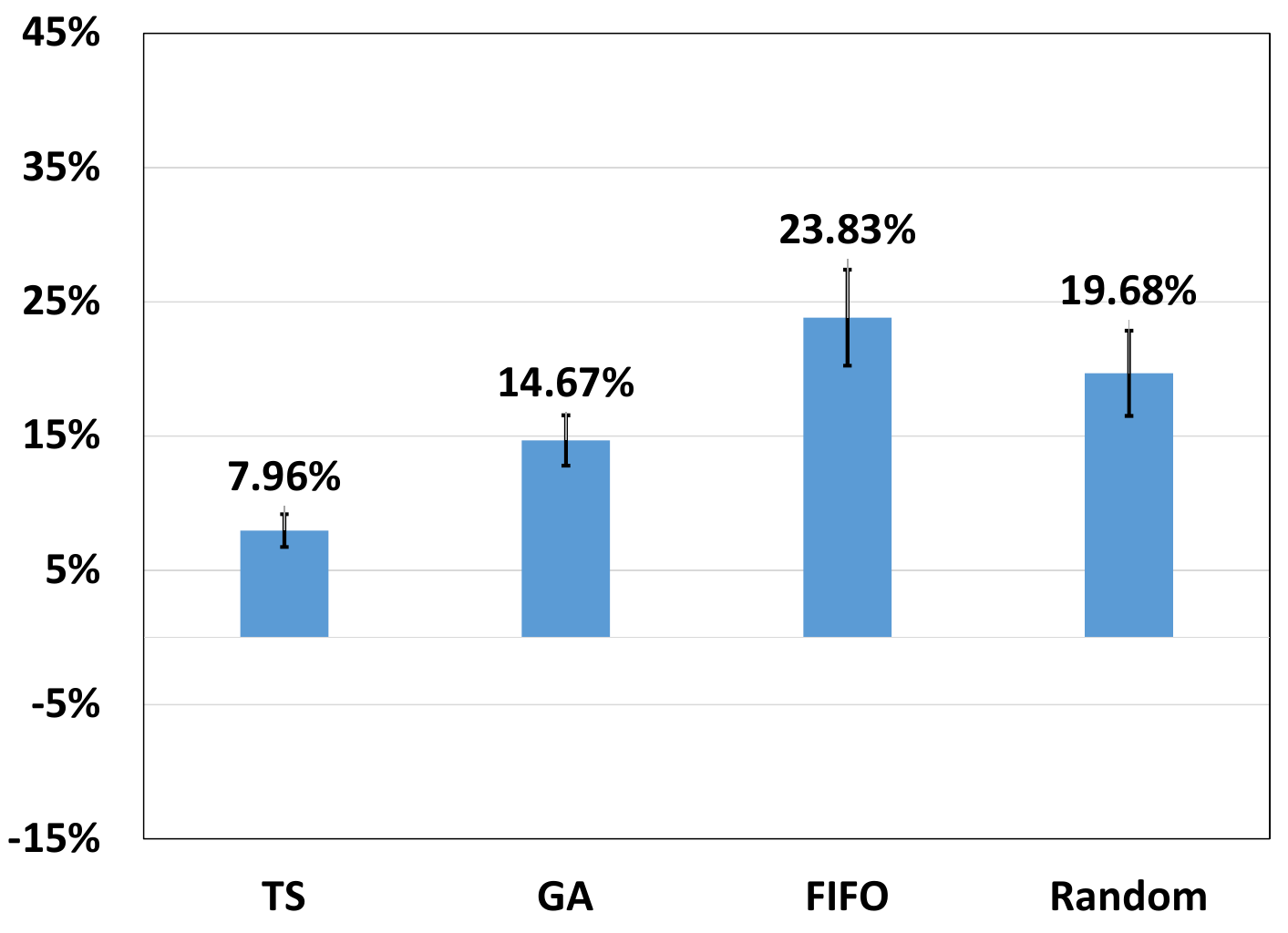}
  \caption{30 Jobs with 4 Machines.}
  \label{fig:all_jobs_ms_compare}
\end{subfigure}\hfil 
\begin{subfigure}{0.5\textwidth}
  \includegraphics[width=\linewidth]{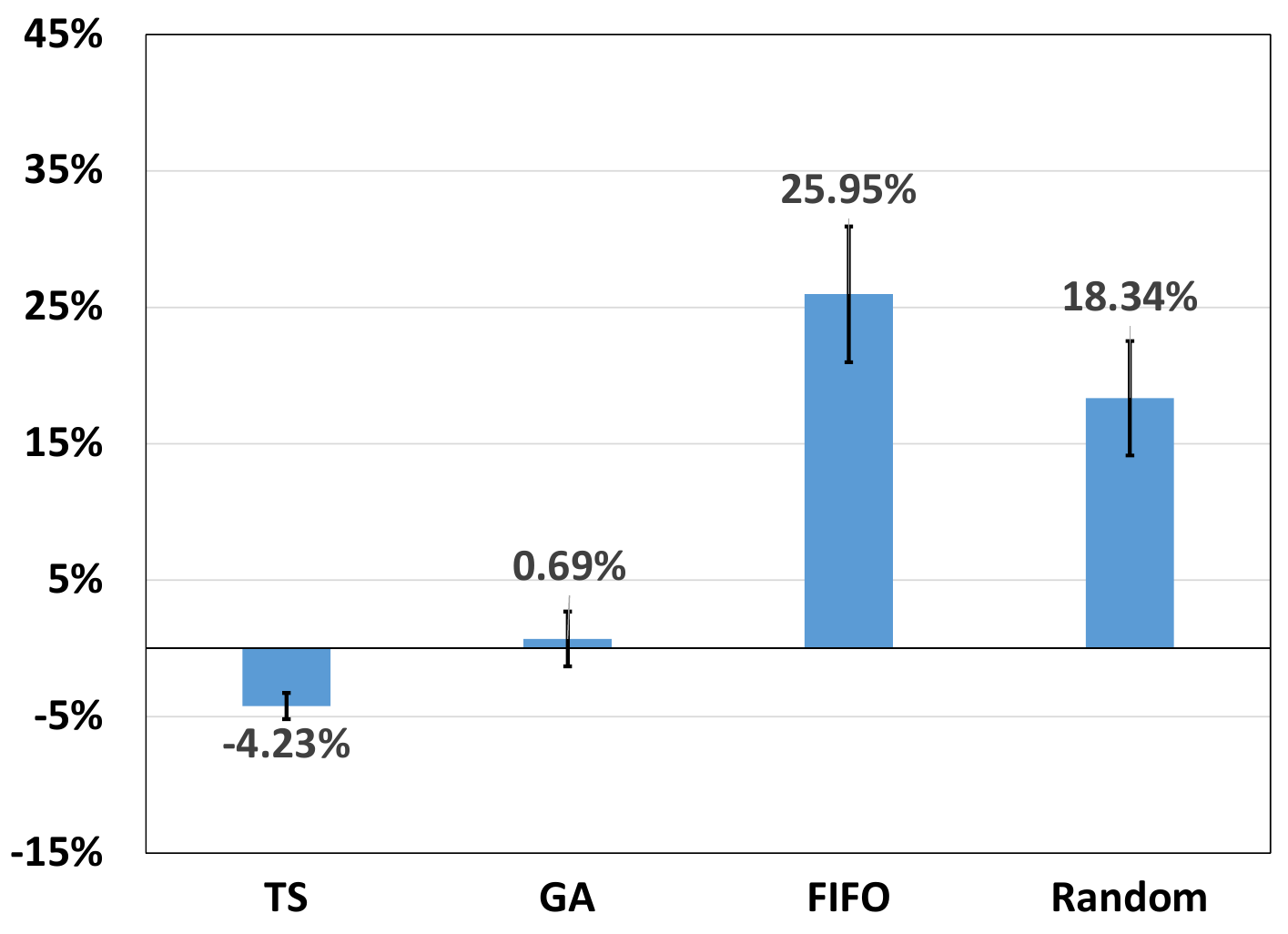}
  \caption{30 Jobs with 8 Machines.}
  \label{fig:double_jobs_ms_compare}
\end{subfigure}\hfil 

\medskip
\begin{subfigure}{0.5\textwidth}
  \includegraphics[width=\linewidth]{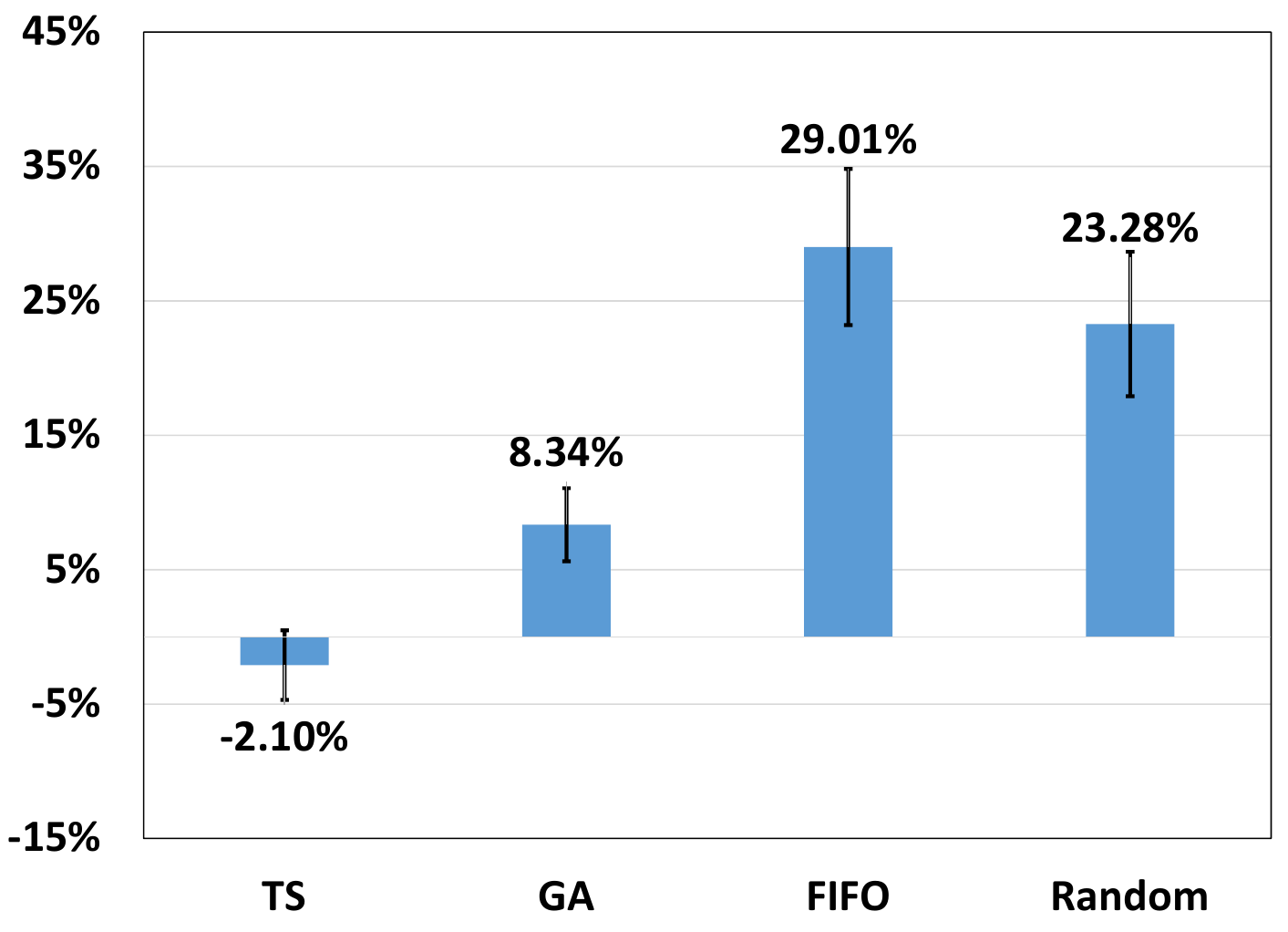}
  \caption{14 Jobs with 4 Machines.}
  \label{fig:2_jobs_ms_compare}
\end{subfigure}\hfil
\begin{subfigure}{0.5\textwidth}
  \includegraphics[width=\linewidth]{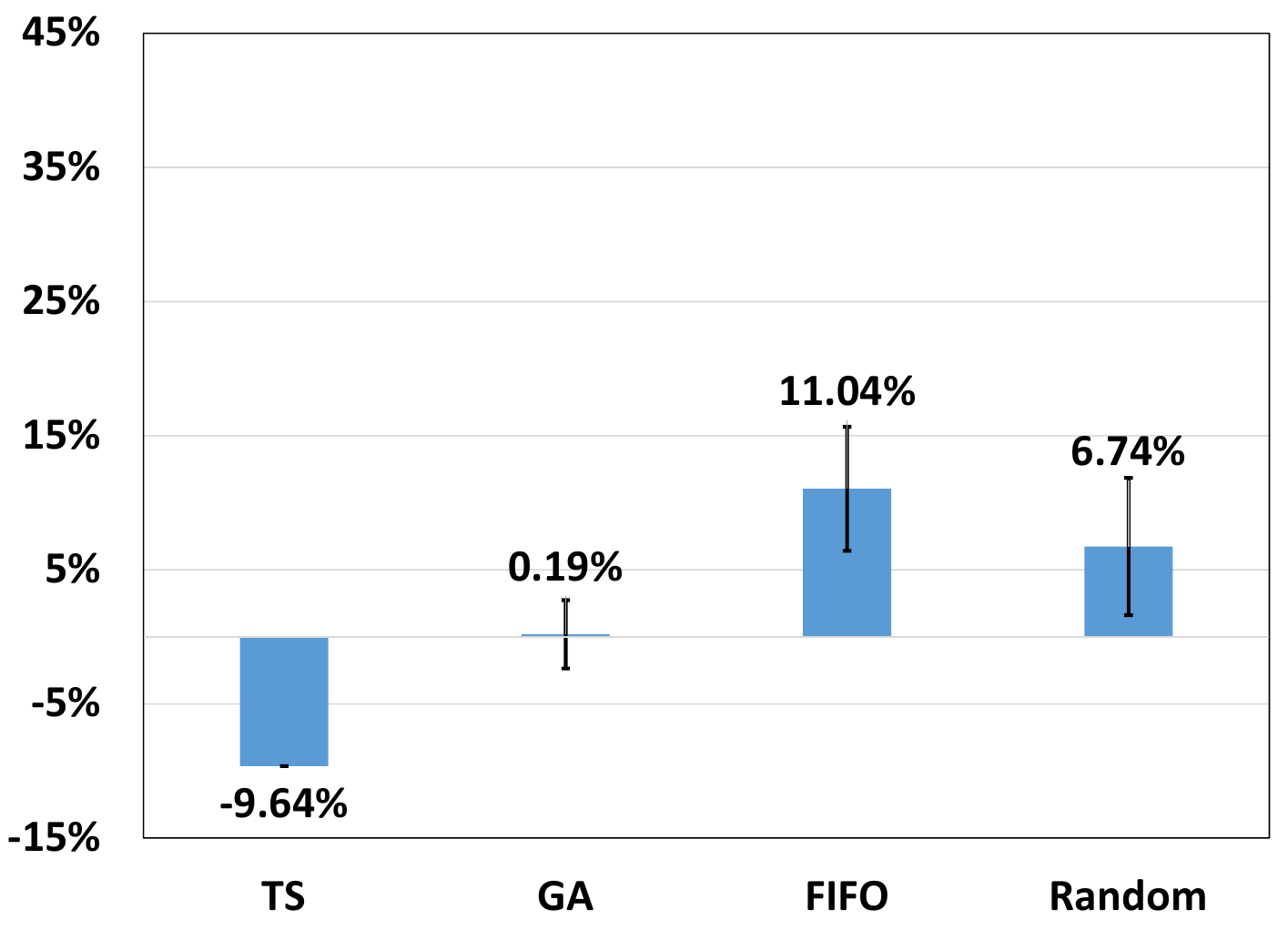}
  \caption{19 Jobs with 4 Machines.}
  \label{fig:3_jobs_ms_compare}
\end{subfigure}\hfil 

\medskip
\begin{subfigure}{0.5\textwidth}
  \includegraphics[width=\linewidth]{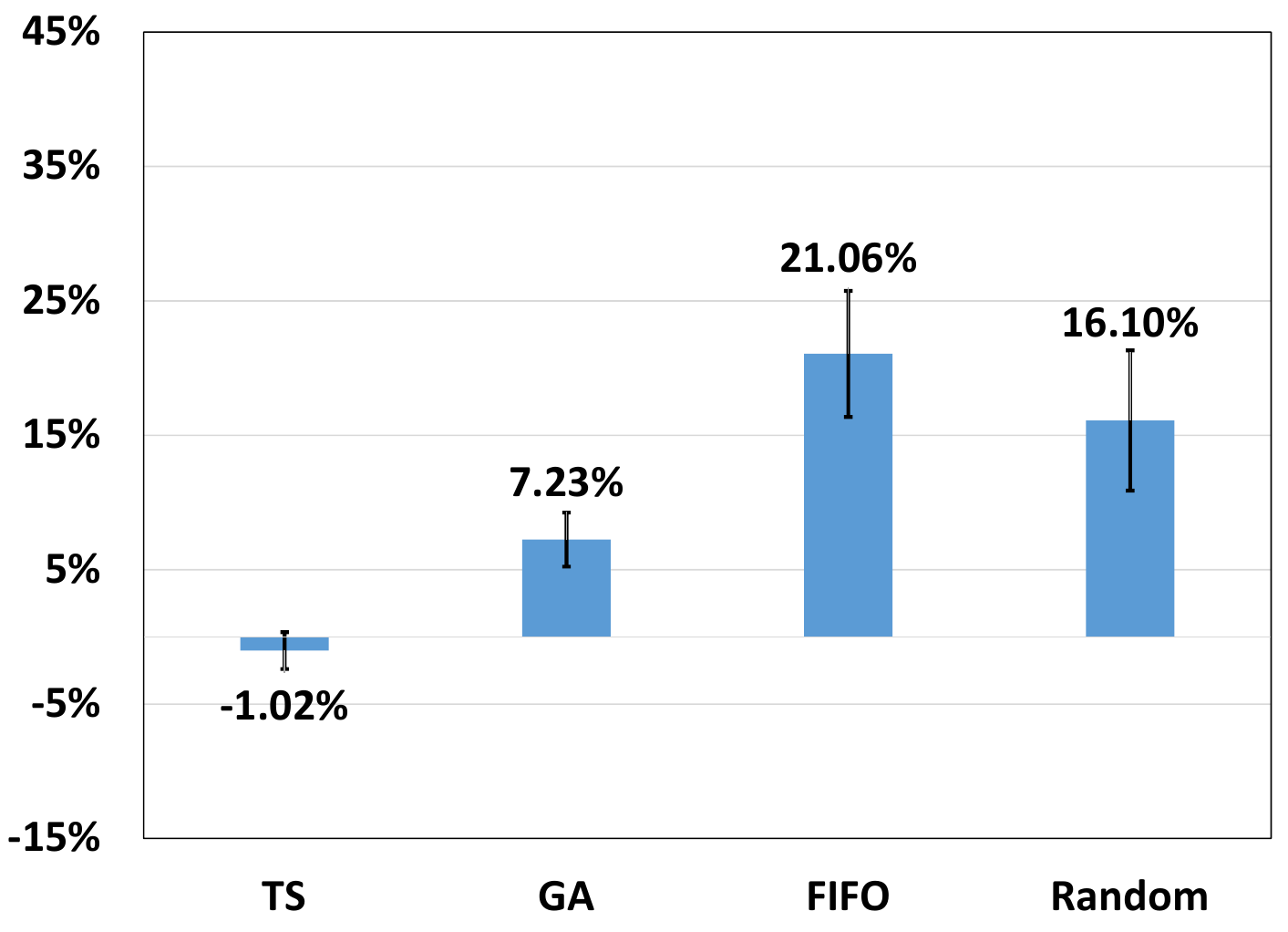}
  \caption{22 Jobs with 4 Machines.}
  \label{fig:4_jobs_ms_compare}
\end{subfigure}\hfil 
\begin{subfigure}{0.5\textwidth}
  \includegraphics[width=\linewidth]{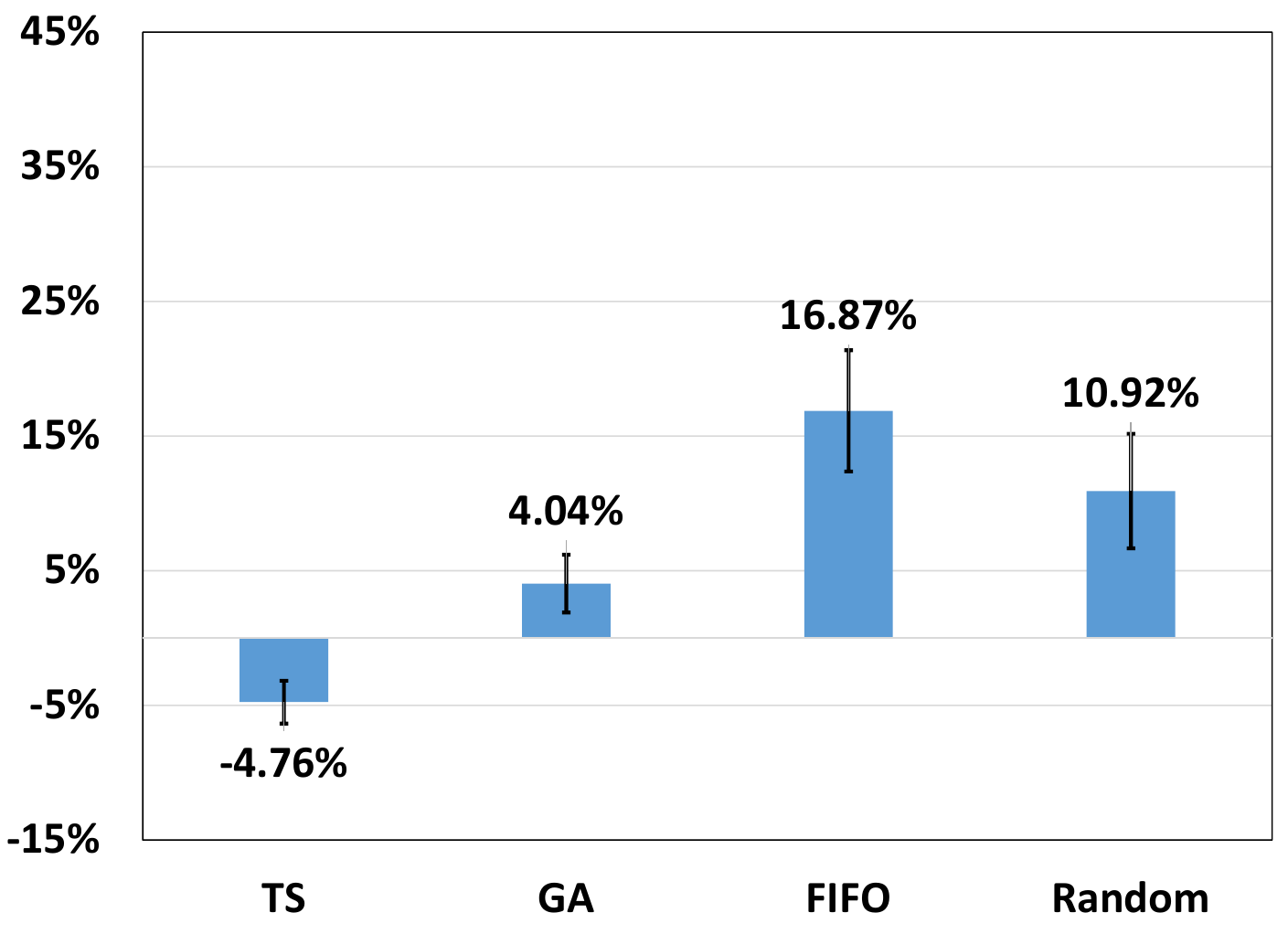}
  \caption{24 Jobs with 4 Machines.}
  \label{fig:5_jobs_ms_compare}
\end{subfigure}\hfil 

\caption{Comparison of the makespan under various conditions of the environment and the jobs. The baseline is the plan made by the GraSP-RL agent, while bars represent the average percentage improvement or deterioration of the corresponding methods compared to the makespan planned by GraSP-RL. The error bars represent the standard deviation in makespan over 100 runs.}
\label{fig:all_makespan_comparison}
\end{figure}

We test the agent in two different problem classes: OSS and RJSSP. In OSS we add one scenario, where the machine capacity is increased. The addition of machines adds complexity to the agent since processing the jobs like before may add additional setup times. So the agent can only perform better if they had understood the future implications of their current actions. And in RJSSP we add four scenarios, where the jobs are removed randomly. The available jobs in the RJSSP perturbations of the environment are as follows: 14, 19, 22, 24. To achieve this, the jobs were split into blocks of sizes 2, 3, and 4, and one of the jobs was randomly removed from the block. Additionally $Job-0$ was omitted to add to the randomness. This initial state was then preserved and further tested with TS, GA, FIFO, and Random actions.

\begin{table}[htp]
\begin{tabular}{lccc}
\hline
                  & JSSP (s)         & OSS (s)          & RJSSP (s)        \\ \hline
GraSP-RL           & 0.51 $\pm$ 0.00  & 0.37 $\pm$ 0.00  & 0.50 $\pm$ 0.00  \\
Tabu Search       & 55.00 $\pm$ 0.14 & 55.26 $\pm$ 0.26 & 58.48 $\pm$ 1.64 \\
Genetic Algorithm & 50.18 $\pm$ 0.01 & 50.18 $\pm$ 0.01 & 50.30 $\pm$ 0.10 \\ \hline
\end{tabular}
\caption{Comparison of the average time taken to make a plan given the input conditions by TS, GA, and GraSP-RL in 100 trials.}
\label{tab:time_comparison}
\end{table}

As shown in Fig.~\ref{fig:all_makespan_comparison}, the RL agent without any further training, performs better than GA, FIFO, and random in all the scenarios, while also being only marginally worse in most cases compared to TS (except in~\ref{fig:3_jobs_ms_compare}). Another huge advantage for the RL agent is that the plan made by it is instantaneous, while the methods like TS and GA take \~100x more time, as shown in Table~\ref{tab:time_comparison}.

\section{Conclusion}\label{sec:conclusion}

JSSP approaches are majorly centralized and they do not scale well with increasing problem size. Even the RL approaches use a vectorized representation of the current state and do not take into consideration the inductive bias present in the graph nature of the JSSP. We overcome this by first setting up the JSSP environment as a bipartite graph, then processing it through a GNN and further using the message passed nodes feature of the machine in RL to take action. And since the RL agent's information is only from the machine's point of view, it learns the local information, relative to its neighboring nodes, to take action on the available jobs in the preceding buffer node. This enables the capture of the relationship between the machines and also enables distributed, decentralized learning directly on the machine unit, while also generalizing well into unseen scenarios.

The effectiveness of the planned solution is then compared with metaheuristic algorithms like TS and GA, and priority dispatch rules like FIFO. As shown by the results, the agent performs well on the JSSP task providing the best plan among the comparison. Further tests on the agent in new problem classes like OSS, and RJSSP show that the agent performs better than FIFO and random actions, while also providing comparable performance to TS and GA without any further training. This is significant for two reasons since the agent can scale to the addition of new machines, therefore eliminating the need for retraining, and the agent can solve problems with perturbations to the initial conditions on which it was trained, which happens frequently in a factory. Furthermore, the agent also provides instantaneous plans, while the metaheuristics take 100x more time. In future work, we plan to improve the encoding of jobs using more sophisticated methods than one-hot encoding.

\bibliography{reference}





\end{document}